\documentclass[final]{cvpr}

\usepackage[T1]{fontenc}
\usepackage{times}
\usepackage{epsfig}
\usepackage{graphicx}
\usepackage{amsmath}
\usepackage{amssymb}
\usepackage{color,soul}     
\usepackage{booktabs}       
\usepackage{caption}        

\usepackage[pagebackref=true,breaklinks=true,colorlinks,bookmarks=false]{hyperref}

\def\cvprPaperID{13} 
\def\confYear{CVPR 2021}

\newcommand{\cgen}{\textbf{CurveGen}}
\newcommand{\tgen}{\textbf{TurtleGen}}
\newcommand{\bez}{B\'ezier}

\iftoggle{cvprfinal}{
    \pagenumbering{gobble}
}

\title{Engineering Sketch Generation for Computer-Aided Design}

\author{
Karl D.D. Willis \quad Pradeep Kumar Jayaraman \quad Joseph G. Lambourne \quad Hang Chu \quad Yewen Pu\\
Autodesk Research\\
}

\begin{document}

\twocolumn[{%
    \renewcommand\twocolumn[1][]{#1}%
    \maketitle
    
    \begin{center}
    \vspace{-0.25cm}
    \centering
    \includegraphics[width=0.92\linewidth]{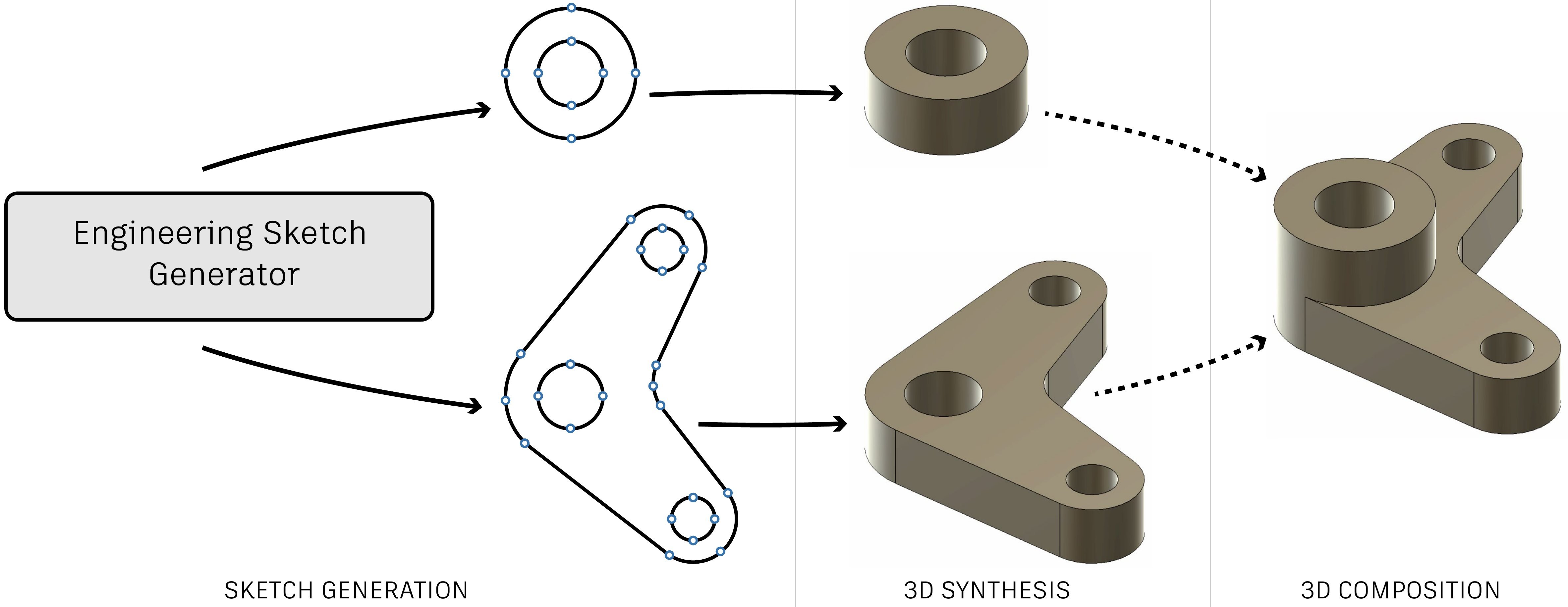}
        \captionof{figure}{We tackle the problem of learning based engineering sketch generation as a first step towards synthesis and composition of solid models with an editable parametric CAD history.} \label{fig:teaser}
    \end{center}
}]

\begin{abstract}
Engineering sketches form the 2D basis of parametric Computer-Aided Design (CAD), the foremost modeling paradigm for manufactured objects. In this paper we tackle the problem of learning based engineering sketch generation as a first step towards synthesis and composition of parametric CAD models. We propose two generative models, CurveGen and TurtleGen, for engineering sketch generation. Both models generate curve primitives without the need for a sketch constraint solver and explicitly consider topology for downstream use with constraints and 3D CAD modeling operations. We find in our perceptual evaluation using human subjects that both CurveGen and TurtleGen produce more realistic engineering sketches when compared with the current state-of-the-art for engineering sketch generation.
\end{abstract}


\section{Introduction}
\label{sec:intro}
Parametric Computer-Aided Design (CAD) is the foremost 3D modeling paradigm used to design manufactured objects from automobile parts, to electronic devices, to furniture. Engineering sketches form the 2D basis of parametric CAD, and refer specifically to composite curves made up of 2D geometric primitives (e.g. lines, arcs, circles), topological information about how the primitives connect together, and constraints defined using topology (e.g. coincidence, tangency, symmetry).
Engineering sketches can be extruded or revolved to generate simple 3D solid bodies (Figure~\ref{fig:teaser}), and in turn combined using Boolean operations to build up complex shapes \cite{MASUDA1993119}. 
This workflow is common to all parametric CAD software and supported by all the major solid modeling kernels.
Consequently, the ability to generate high quality engineering sketches is a major enabling technology for the automatic generation of solid models with an editable parametric CAD history. 

Engineering sketch generation can be applied in a number of CAD workflows. For example, a long sought-after goal is the ability to automatically reverse engineer a parametric CAD model from noisy 3D scan data \cite{BENKOXB2002173}. One way to realize this goal is to generate 2D CAD sketches from sparse scan data, just like human designers would, and apply suitable modeling operations to reconstruct the 3D CAD model. 
Engineering sketch generation can also be applied to auto-completion of user input. The ability to infer repetitive commands based on visual or geometric input could significantly ease user burden when producing complex engineering sketches. Another sought-after capability is the generation of engineering sketches from approximate geometry like free-hand drawings. Often referred to as \emph{beautification}, a generative model for engineering sketches could potentially improve user workflows over traditional approaches~\cite{FISER201646}.  

Despite recent advances with 2D vector graphic generation using data-driven approaches \cite{lopes2019learned, carlier2020deepsvg, reddy2021im2vec}, there exists limited research on synthesizing engineering sketches directly. This is a challenging problem because engineering sketches contain disparate 2D geometric primitives coupled with topological information on how these primitives are connected together. The topology information is critical to ensure: 1) geometric primitives can be grouped into closed profile loops and lifted to 3D with modeling operations, and 2) constraints can be correctly defined using the topology, for example, two lines can be constrained to intersect at 90$^\circ$ if their endpoints are known to coincide. With the availability of large-scale engineering sketch and parametric CAD datasets~\cite{seff2020sketchgraphs,willis2020fusion}, we believe a data-driven approach that learns a generative model is a promising avenue to explore.

In this paper, we propose two generative models, \textbf{CurveGen} and \textbf{TurtleGen}, for the task of engineering sketch generation. CurveGen is an adaptation of PolyGen~\cite{nash2020polygen}, where Transformer networks are used to predict the sketches autoregressively. While PolyGen generates polygonal mesh vertices and indexed face sets, CurveGen generates vertices lying on individual curves and indexed hyperedge sets that group the vertices into different curve primitives (line, arc, circle). This approach generates not only the individual curve geometry, but also the topology linking different curves, a vital requirement of engineering sketches. 
TurtleGen is an extension of TurtleGraphics~\cite{goldman2004turtle}, where an autoregressive neural network generates a sequence of \emph{pen\_down, pen\_draw, pen\_up} commands, creating a series of closed loops which forms the engineering sketch. 
We find in our perceptual evaluation using human subjects that both CurveGen and TurtleGen produce more realistic engineering sketches when compared with the current state-of-the-art for engineering sketch generation.
Importantly, both CurveGen and TurtleGen can generate geometry and topology directly without the expense of using a sketch constraint solver.
This paper makes the following contributions:

\begin{itemize}
    \item We introduce two generative models which tackle the problem of engineering sketch generation without the use of a sketch constraint solver.
    \item We use a novel sketch representation with our CurveGen model that implicitly encodes the sketch primitive type based on hyperedge cardinality.
    \item We show quantitative and qualitative results for engineering sketch generation, including the results of a perceptual study comparing our generative models with a  state-of-the-art baseline.
\end{itemize}

\section{Related Work}
\label{sec:related}
With the advent of deep learning a vast body of work has focused on generation of novel raster images.
By contrast, significantly fewer works have tackled image generation using parametric curves such as lines, arcs, or \bez{} curves. Parametric curves are widely used in 2D applications including vector graphics, technical drawings, floor plans, and engineering sketches---the focus of this paper. 
In this section we review work related to engineering sketch generation and its application to the synthesis and composition of solid CAD models.

\paragraph{Vector Graphics}
Vector graphics are used extensively in commercial software to enable the resolution independent design of fonts, logos, animations, and illustrations.
Although a rich body of work has focused on \textit{freeform} sketches~\cite{xu2020deep, chuan2020sketchcleanupbench}, we narrow our focus to structured vector graphics that consider shape topology, connectivity, or hierarchy relevant to engineering sketches. 

Fonts are described with curves that form closed loop profiles and can contain interior holes to represent letters of different genus (e.g. \textit{b} or \textit{B}). SVG-VAE~\cite{lopes2019learned} is the first work to learn a latent representation of fonts using a sequential generative model for vector graphic output. DeepSVG~\cite{carlier2020deepsvg} considers the structured nature of both fonts and icons with a hierarchical Transformer architecture.
\bez{}GAN~\cite{chen2020beziergan} synthesizes smooth curves using a generative adversarial network~\cite{goodfellow2014generative} and applies it to 2D airfoil profiles.
More recently, Im2Vec~\cite{reddy2021im2vec} leverages differentiable rendering~\cite{li2020differentiable} for vector graphic generation trained only on raster images. 

Common in prior research is the use of \bez{} curves. Line, arc, and circle primitives are preferred in engineering sketches as they are easier to control parametrically, less prone to numerical issues (e.g. when computing intersections, offsetting, etc.) and can be lifted to 3D prismatic shapes like planes and cylinders rather than NURBS surfaces. In contrast to prior work, we pursue engineering sketch generation with line, arc, and circle primitives and explicitly consider topology for downstream application of constraints and use with 3D CAD modeling operations.

\paragraph{Technical Drawings and Layout}
Technical drawings take the form of 2D projections of 3D CAD models, often with important details marked by dimensions, notes, or section views. 
Han \etal~\cite{Han2020SPARE3D} present the SPARE3D dataset, a collection of technical drawings generated from 3D objects with three axis-aligned and one isometric projection.  The dataset is primarily aimed at spatial reasoning tasks.
Pu \etal~ \cite{Pu2005retrieval} show how freehand 2D sketches can be used to assist in the retrieval of 3D models from large object databases.  Their system allows designers to make freehand sketches approximating the engineering sketches which would be used in the construction of a 3D model.  
Egiazarian \etal~\cite{egiazarian2020deep} address the long standing problem of image vectorization for technical drawings. They predict the parameters of lines and \bez{} curves with a Transformer based network, then refine them using optimization. In contrast, our system focuses on the generation of new sketch geometry suitable for use with 3D CAD modeling operations.

An emerging area of research considers learning based approaches to generative layout for graphic design~\cite{li2019layoutgan,zheng2019content,lee2020neural} and floor plans~\cite{nauata2020house, hu2020graph2plan}. Although a different domain than engineering sketches, these works output high level primitives (e.g. rooms in a floor plan) that must be correctly connected to the overall layout and obey relevant constraints (e.g. number of bedrooms).
Layout problems have similarities to engineering sketches where connectivity between parts of the sketch and relationships, such as parallel, symmetric, or perpendicular curves, are critical when designing.

\paragraph{Program Synthesis}
Another potential approach to engineering sketch generation is program synthesis. Recent work in this domain leverages neural networks in combination with techniques from the programming language literature to generate or infer programs capable of representing 2D~\cite{ellis2017learning, ellis2020dreamcoder, sharma2018csgnet} or 3D~\cite{sharma2018csgnet, tian2018learning, ellis2019write, kania2020ucsg} geometry. 
One can view engineering sketch generation as a program synthesis task where the geometry is represented as a sequence of programmatic commands that draw vertices and curves, constructing the sketch one piece at a time. Future applications of program synthesis would allow us to model more complex operations with programmatic constructs, such as repetitions (loops), symmetries (constraints), and modifications (refactoring and debugging).

\paragraph{Reverse engineering}
Another important field is reverse engineering 3D models from scan data or triangle meshes \cite{Buonamici2018}. Typically 2D poly-linear profiles are generated by intersecting the mesh data with a plane, and then line and arc primitives can be least squares fitted in a way which respects a series of constraints \cite{BENKOXB2002173}. Generative models which can be conditioned on approximate data provide an approach towards automating this part of the reverse engineering procedure.
Recent learning-based approaches have tackled the challenge of reverse engineering parametric curves~\cite{gao2019deepspline, wang2020pie}, paving the way for reconstruction of trimmed surface models~\cite{li2019supervised,smirnov2019learning,smirnov2020deep,sharma2020parsenet}. In contrast to these works, we focus on 2D sketch generation as a building block towards the synthesis and composition of solid models, with parametric CAD history, using common modeling operations.

\paragraph{Engineering Sketches}

Most closely related to our work is SketchGraphs~\cite{seff2020sketchgraphs}, a recently released dataset and baseline generative model for the task of engineering sketch generation. The SketchGraphs dataset consists of engineering sketches made by users of the Onshape CAD software. The SketchGraphs generative model works by predicting the underlying geometric constraint graph, and relies on a sketch constraint solver to determine the final configuration of the sketch. Unlike SketchGraphs we directly predict geometry and are not reliant on a sketch constraint solver. We observe qualitatively that our methods naturally generate geometry which conforms to the regular patterns seen in engineering sketches, such as horizontal and vertical lines and symmetries. The network can be considered to encode the constraint information \emph{implicitly} in the geometric coordinates, allowing constrained sketches to be recovered in a post-processing step if required.  We compare our work directly to SketchGraphs and present results that show we are able to produce more realistic engineering sketch output.

\section{Method}
\label{sec:method}

\subsection{Engineering Sketch Representation}
To create an engineering sketch representation we consider a number of factors.
1) Engineering sketches are composed primarily from lines, arcs, and circles, while ellipses and splines are used less frequently~\cite{seff2020sketchgraphs,willis2020fusion}. 2) Engineering sketches must obey constraints such as forming closed profiles, coinciding the end points, and forming 90 degree angles. 3) As the constraints remove many degrees of freedom, engineering sketches are both structured and sparse when compared to free-form sketches or vector graphics. We present two representations of engineering sketches that account for these design considerations: Sketch Hypergraph representation (used by CurveGen) and Turtle Graphics representation (used by TurtleGen).

\begin{figure}
    \centering
    \includegraphics[width=1\linewidth]{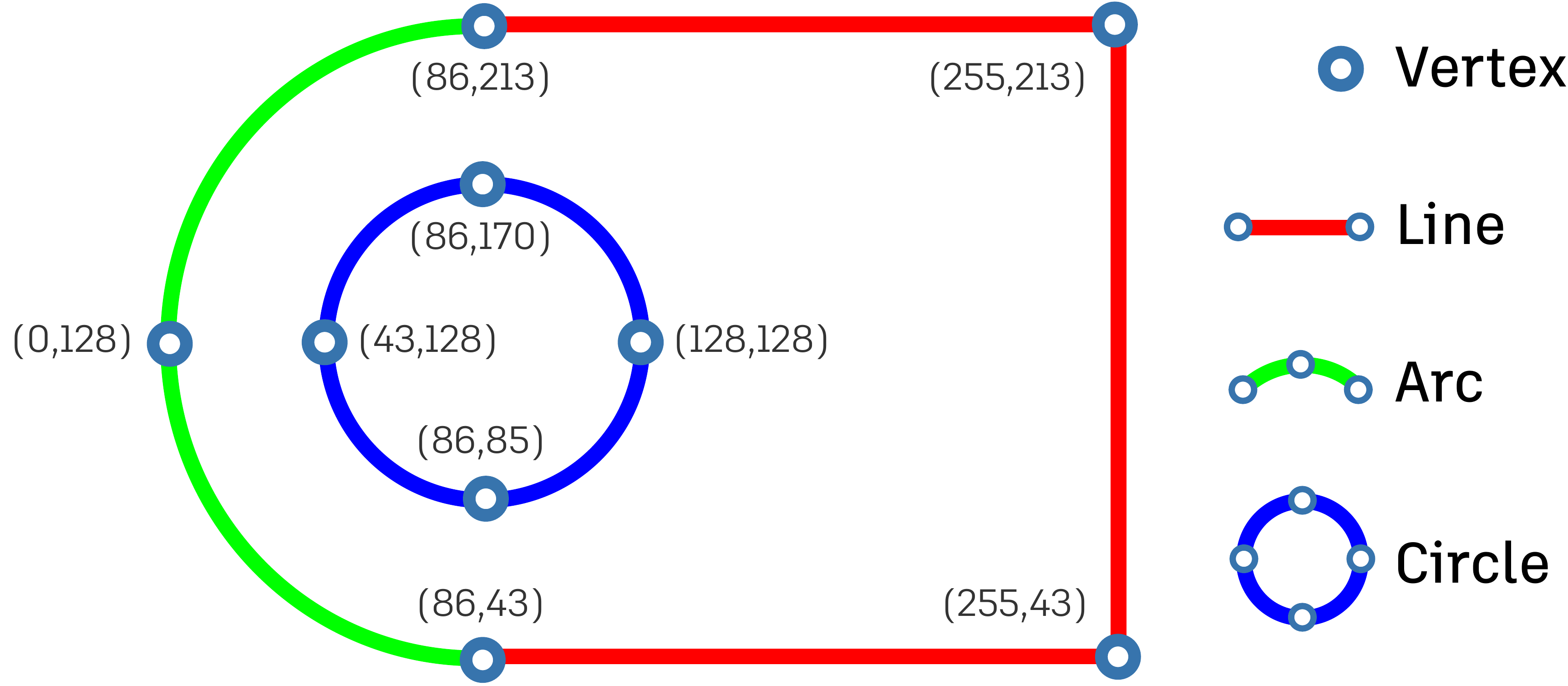} 
    \caption{A hypergraph representation of a sketch consisting of three lines (2 vertices), an arc (3 vertices), and a circle (4 vertices).}
    \label{fig:sketch_rep}
\end{figure}

\paragraph{Sketch Hypergraph Representation}
\label{sec:sketch_rep_hypergraph}

Under this representation, a sketch is represented as a hypergraph $G=(V,E)$ where $V = \{v_1, v_2, \ldots, v_n\}$ denotes the set of $n$ vertices, each vertex $v_i = (x_i,y_i)$ consists of a vertex id $i$ and its corresponding location $x_i, y_i$. $E$ denotes a set of hyperedges that connects 2 or more vertices to form geometric primitives. The primitive curve type is implicitly defined by the cardinality of the hyperedge: line (2), arc (3), circle (4).
Arcs are recovered by finding the circle that uniquely passes through the 3 vertices.
Circles are recovered by a least squares fitting due to the over-parameterization by 4 points.
Figure \ref{fig:sketch_rep} shows the sketch hypergraph representation, consisting of 9 vertices and 5 hyperedges. In addition, the vertices are quantized to a 256$\times$256 grid to bias the generative model into generating few distinct coordinates by learning to produce repeated $(x, y)$ values. 

\begin{figure*}
    \centering
    \includegraphics[width=1\linewidth]{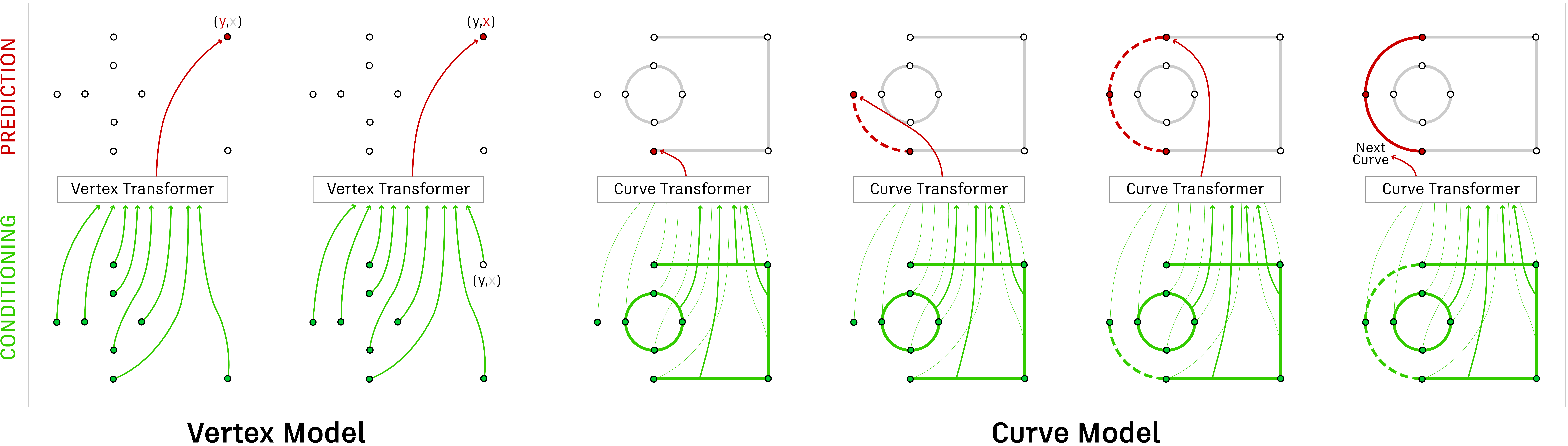} 
    \caption{CurveGen, our variant of the PolyGen~\cite{nash2020polygen} architecture, first generates sketch vertices (left), and then generates sketch curves conditioned on the vertices (right).}
    \label{fig:network_curve_gen}
\end{figure*}

\paragraph{Turtle Graphics Representation}
The Turtle Graphics representation uses a \emph{sequence} of drawing commands, which can be \emph{executed} to form an engineering sketch in the hypergraph representation. Intuitively, the turtle representation can be thought of as a sequence of pen-up, pen-move actions, which iteratively \emph{draws} the engineering sketch one loop at a time. It is used by Ellis \etal \cite{ellis2020dreamcoder} to generate compositional and symmetric sketches. Specifically, the drawing commands of turtle graphics are specified by the following grammar:
\[
\begin{array}{rcl}
Turtle &\vdash& [~Loop~] \\
Loop &\vdash& LoopStart ~[Draw] \\
Draw &\vdash& Line ~|~ Arc ~|~ Circle \\
LoopStart &\vdash& loopstart (\Delta) \\
Line &\vdash& line (\Delta) \\
Arc &\vdash& arc (\Delta,\Delta) \\
Circle &\vdash& circle (\Delta,\Delta,\Delta,\Delta) \\
\Delta &\vdash& (int, int) \\
\end{array}
\]

Here, a $Turtle$ program consists of a sequence of $Loop$s, each loop consists of a start command $LoopStart$, followed by a sequence of $[Draw]$ commands. The pen initially starts at $(0,0)$. The $LoopStart$ command starts a new loop by lifting the current pen, displacing/teleporting it by $\Delta$ and putting it back down. The $Draw$ command can be one of three primitives $Line$, $Arc$, and $Circle$, each parameterized by a different number of $\Delta$ displacements, which extend the current loop without lifting the pen, displacing it \emph{relative} to the current pen location. After a loop is completed, the pen returns/teleports back to $(0,0)$. As with the hypergraph representation,  $\Delta$ values are quantized to a 256$\times$256 grid. The loops are ordered so that ones closest to $(0,0)$, where distance is measured between the loop's closest vertex to $(0,0)$, are drawn first. We provide an example program in Section~\ref{sec:sup_turtle_program} of the Supplementary Material.

\subsection{Generative Models}
We design and compare two different neural architectures on the task of engineering sketch generation: \cgen{} and  \tgen{}.

\subsubsection{CurveGen}
CurveGen is our adaptation of the PolyGen~\cite{nash2020polygen} architecture applied to engineering sketch generation. CurveGen generates the sketch hypergraph representation \textit{directly}.
As with the original PolyGen implementation, we break the generation of $G$ into two steps based on the chain rule: 1) generate the sketch vertices $V$, 2) generate the sketch hyperedges $E$ conditioned on the vertices:
\[
p(G) = \underbrace{p(E|V)}_{\substack{\text{Curve}\\ \text{model}}} \underbrace{p(V)}_{\substack{\text{Vertex}\\ \text{model}}} \ ,
\]
where $p(\cdot)$ are probability distributions.
Figure~\ref{fig:network_curve_gen} illustrates the two step generation process starting with the vertex model (left) and then the curve model (right). The final stage of recovering the curve primitives from the hyperedges is done as a post-process.
We use the vertex model directly from PolyGen with 2D vertex coordinates and adapt our curve model from the PolyGen face model to work with 2D curves.
We use 3 Transformer blocks in the vertex decoder, curve encoder, and curve decoder Transformer models~\cite{vaswani2017attention}.
Each block includes a multihead attention with 8 heads, layer normalization and a 2-layer multilayer perceptron (MLP) with 512D hidden, and 128D output dimensions.
Dropout with rates 0.2 and 0.5 are applied in each block of the vertex and curve models, respectively, right after the MLP.
The vertex and curve models are both trained by negative log likelihood loss against the ground truth data. Once the neural network is trained, we perform nucleus sampling~\cite{holtzman2019curious} to directly generate samples in the hypergraph sketch representation.
We refer the reader to Nash \etal~\cite{nash2020polygen} for further details.

\subsubsection{TurtleGen}
The neural network for generating a program in the Turtle representation is a sequence generator. The sequence of turtle commands are encoded as a sequence of discrete valued tokens, where each of the commands \textit{loopstart}, \textit{line}, \textit{arc}, \textit{circle}, along with two tokens for sequence \textit{start} and \textit{end}, are represented as 1-hot vectors. The quantized integer coordinates are encoded as two 1-hot vectors for \textit{x} and \textit{y} each. For any given sketch in the hypergraph representation, we randomize the turtle sequence generation by randomly selecting the loop order, loop's starting vertex, and the direction of drawing the loop. We discard long sequences over 100 turtle commands.

The neural network is a simple 9-layer Transformer with 512D hidden and 128D output dimensions. The network has seven linear branches of input and output, where the first branch corresponds to the type of command, and the remaining six branches correspond to three \textit{x} and \textit{y} coordinates. Commands with less than three points are padded with zeros.
The input branches are concatenated and added with the conventional positional embedding after a linear layer, before they are fed into the Transformer network. The output branches are connected to the Transformer encoding at the previous sequence step. We store the first three ground-truth turtle steps from the training set as a fixed dictionary, which we randomly sample from to precondition the auto-regressive model at sampling time. Each sampled sequence is then executed to recover the sketch hypergraph representation, and define the geometry and topology.

\section{Results}
\label{sec:results}

We now present quantitative and qualitative results on the task of engineering sketch generation, comparing the CurveGen and TurtleGen generative models with the SketchGraphs~\cite{seff2020sketchgraphs} generative model.

\subsection{Data Preparation}
\label{sec:data-prep}
We use the pre-filtered version of the SketchGraphs~\cite{seff2020sketchgraphs} dataset that contains 9.8 million engineering sketches with 16 or fewer curves. We remove duplicates from the dataset to promote data diversity during training, and ensure evaluation on unseen data at test time. We consider two sketches to be duplicates if they have identical topology and similar geometry.
To detect similar geometry, sketches are uniformly scaled and the coordinates are quantized into a 9$\times$9 grid.  Vertices are considered identical if they lie in the same grid square.  The same quantization is also applied to the radii of circles and arcs.  Lines are considered identical if the end points are identical.  Arcs additionally check for an identical quantized radius.  Circles check the center point and radius. We do not consider a sketch unique if both the topology and geometry match, but curve or vertex order is different.
Using this approach we find that duplicates make up 87.01\% of the SketchGraphs data.
We remove all duplicates, invalid data (e.g. sketches with only points) and omit construction curves. We use the official training split and after filtering have 1,106,328 / 39,407 / 39,147 sketches for the train, validation, and test sets respectively.

\begin{table*}
    \centering
    \caption{Quantitative sketch generation results. \textit{Bits per Vertex} and \textit{Bits per Sketch} are the negative log-likelihood calculated over the test set; both are not directly comparable and reported separately for the vertex/curve models used in CurveGen. \textit{Unique}, \textit{Valid}, and \textit{Novel} are calculated over 1000 generated sketches.
    }
    \small
    \begin{tabular}{rcccccc}
    \toprule
    \textbf{Model} & \textbf{Parameters} & \textbf{Bits per Vertex}     & \textbf{Bits per Sketch} & \textbf{Unique \%} & \textbf{Valid \%} & \textbf{Novel \%} \\
    \midrule
    CurveGen       & \textbf{2,155,542}           & 1.75~/~0.20                & 176.69~/~30.64         & \textbf{99.90}             & \textbf{81.50}            & \textbf{90.90} \\
    TurtleGen      & 2,690,310          & 2.27                        & 54.54                   & 86.40             & 42.90           & 80.60 \\    
    SketchGraphs   &        18,621,560             & -                            & 99.38                        & 76.20             & 65.80            & 69.10\\
    SketchGraphs (w/ Duplicates)  &   18,621,560   & -                            & 94.30                   & 58.70                  & 74.00                 &  49.70\\    
    \bottomrule
    \end{tabular}
    \label{tab:sketch_gen_quant}
\end{table*}

\subsection{Experiment Setup}
We train the CurveGen model on the SketchGraphs training set. Unlike the ShapeNet dataset used to train the original Polygen model, the SketchGraphs data does not have any notion of classes and represents a more challenging, but realistic scenario where human annotated labels are unavailable. We train for 2 million iterations with a batch size of 8 using 4 Nvidia V100 GPUs. We use the Adam optimizer with learning rate 5e-4, and apply to the curve model data a jitter augmentation with the amount sampled from a truncated normal distribution with mean 0, variance 0.1 and truncated by the bounding box of the sketch vertices. Training time takes approximately 48 hours. We save the model with the lowest overall validation loss for evaluation.

We train the TurtleGen model on the same data. To ensure fair comparison, we train the model with a batch size of 128 for a total of 0.5 million iterations, which exposes the model to the same number of training data samples as CurveGen. Training is done with a single Nvidia V100 GPU and takes around 48 hours. We use the Adam optimizer with a learning rate of 5e-4, as well as a learning rate scheduler that decreases the learning rate by a factor of 0.5 when the validation loss plateaus. Validation is conducted once every 500 training iterations. We save the model with the lowest overall validation loss for evaluation.

We train the SketchGraphs generative model using the official implementation with and without duplicates.  We train for 150 epochs, as in the original paper, on a single  Quadro RTX 6000 GPU. Training time takes approximately 27 hours.   Following the advice of the SketchGraphs authors, we adjust the learning rate scheduler to reduce the learning rate at epochs 50 and 100.  All other hyperparameters and settings follow the official implementation, including the prediction of numerical node features.  These provide improved geometry initialization before the data is passed to the OnShape constraint solver.

\subsection{Quantitative Results}
\label{sec:quant-results}

\subsubsection{Metrics}
For quantitative evaluation we report the following metrics. \textbf{Bits per Vertex} is the negative log-likelihood of test examples averaged per-vertex and converted from nats to bits; lower is better.
\textbf{Bits per Sketch} is the negative log-likelihood of test examples averaged per-sketch as a whole in bits; lower is better. For CurveGen we report the bits for both the vertex and curve models.
\textbf{Unique} is the percentage of \textit{unique} sketches generated within the sample set. We use the duplicate detection method described in Section~\ref{sec:data-prep} to find the percentage of unique sketches. A lower value indicates the model outputs more duplicate sketches. 
\textbf{Valid} is the percentage of \textit{valid} sketches generated. Invalid sketches include curve fitting failures, curves generated with \textgreater4 vertices, or identical vertices within a curve.
\textbf{Novel} is the percentage of \textit{novel} sketches generated that are \textit{not} identical to sketches in the training set. We again use the duplicate detection method described in Section~\ref{sec:data-prep}.
We evaluate the bits per vertex/sketch metrics on the withheld test set. All other metrics are evaluated on 1000 generated samples. We include non-\textit{Valid} sketches, which often contain valid curves, when calculating the \textit{Unique} and \textit{Novel} metrics.

\subsubsection{Quantitative Comparison}
Table~\ref{tab:sketch_gen_quant} shows the results comparing CurveGen and TurtleGen with the SketchGraphs generative model. The parameter count for each model is provided for reference. Due to differences in sketch representation and terms in the negative log likelihood loss, the bits per vertex and bits per sketch results are not directly comparable between models, and only provided here for reference.  For the SketchGraphs model, the prediction of numerical node features adds an additional term into the loss, giving a higher bits per sketch value than reported in the SketchGraphs paper.
Invalid sketches occur most frequently with TurtleGen, where identical vertices within a curve are often predicted. Invalid sketches from SketchGraphs are commonly due to arcs of near zero length.
For the novel metric, it is reasonable to expect generative models to produce some identical sketches to those in the training data, such as simple circles, rectangles, and combinations thereof. 
The low percentage of novel sketches generated by SketchGraphs when trained on the dataset \textit{with duplicates} suggests that the model memorizes sketches which are duplicated in the training data.  Removing duplicates from the data helps improve variety in the output. For the remainder of the paper we report results from all models trained without duplicates.

\begin{figure}[b]
    \centering
    \includegraphics[width=1\linewidth]{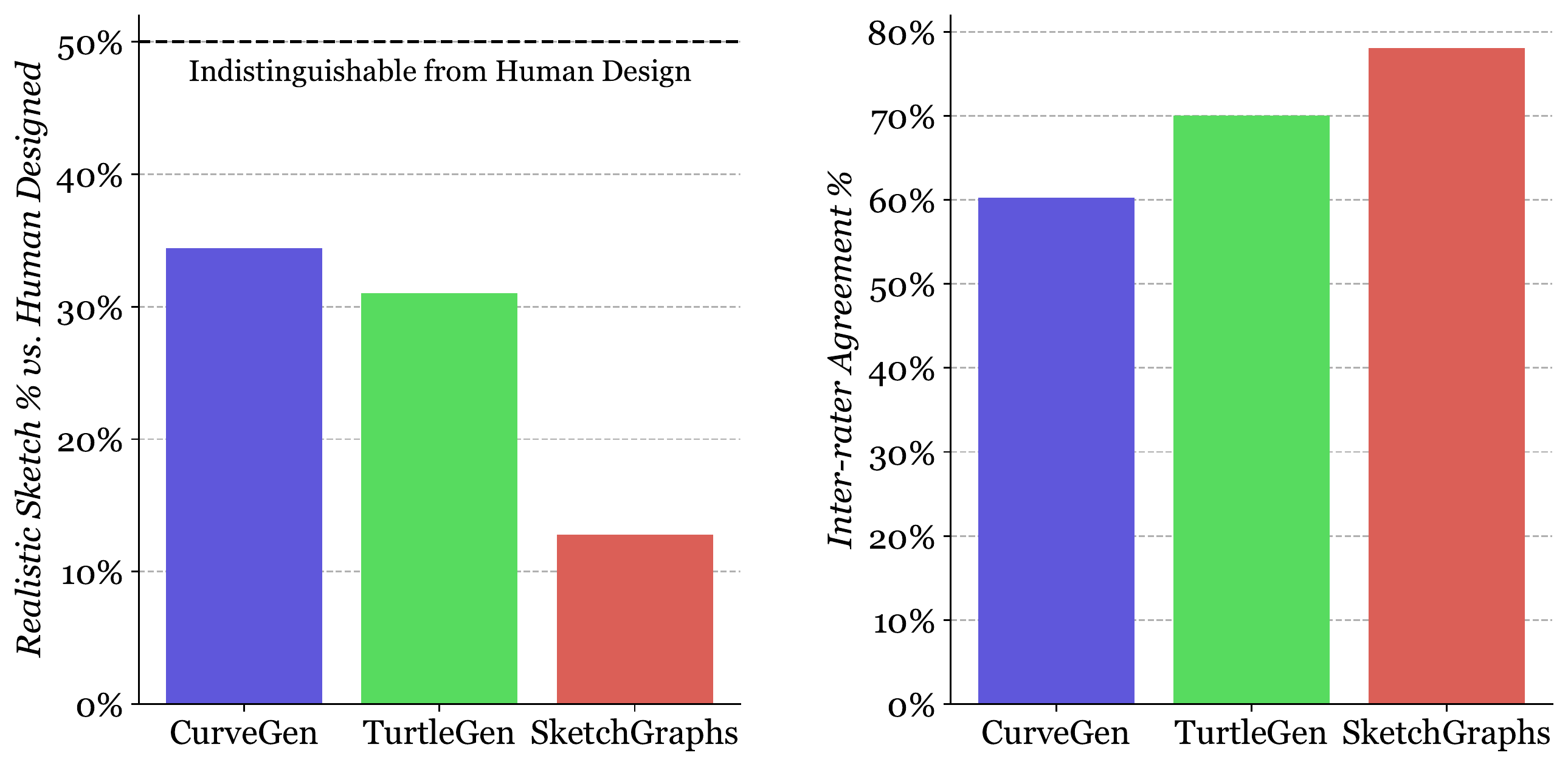} 
    \caption{Results from our perceptual evaluation using human subjects to identify the most realistic engineering sketch.
        Left: The percentage of generated sketches classed as more realistic than human designed sketches.
        Right: The percentage of inter-rater agreement.}
    \label{fig:study_result}
\end{figure}

\begin{figure*}
    \centering
    \includegraphics[width=0.98\linewidth]{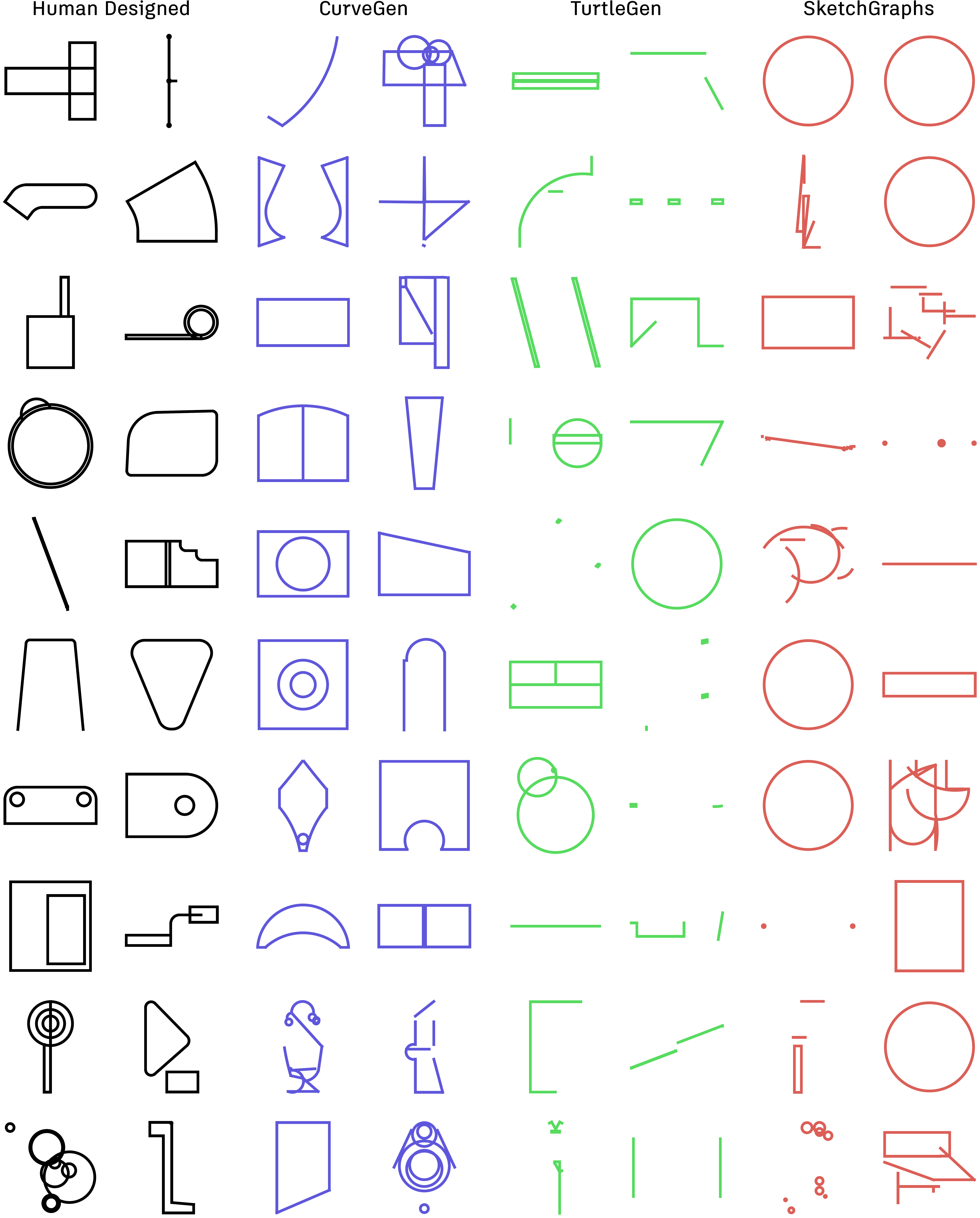} 
    \caption{Qualitative sketch generation results. From left to right: human designed sketches from the SketchGraphs dataset, randomly selected sketches generated using the CurveGen, TurtleGen, and SketchGraphs generative models.}
    \label{fig:sketch_gen_qual}
\end{figure*}

\begin{figure*}
    \centering
    \includegraphics[width=1\linewidth]{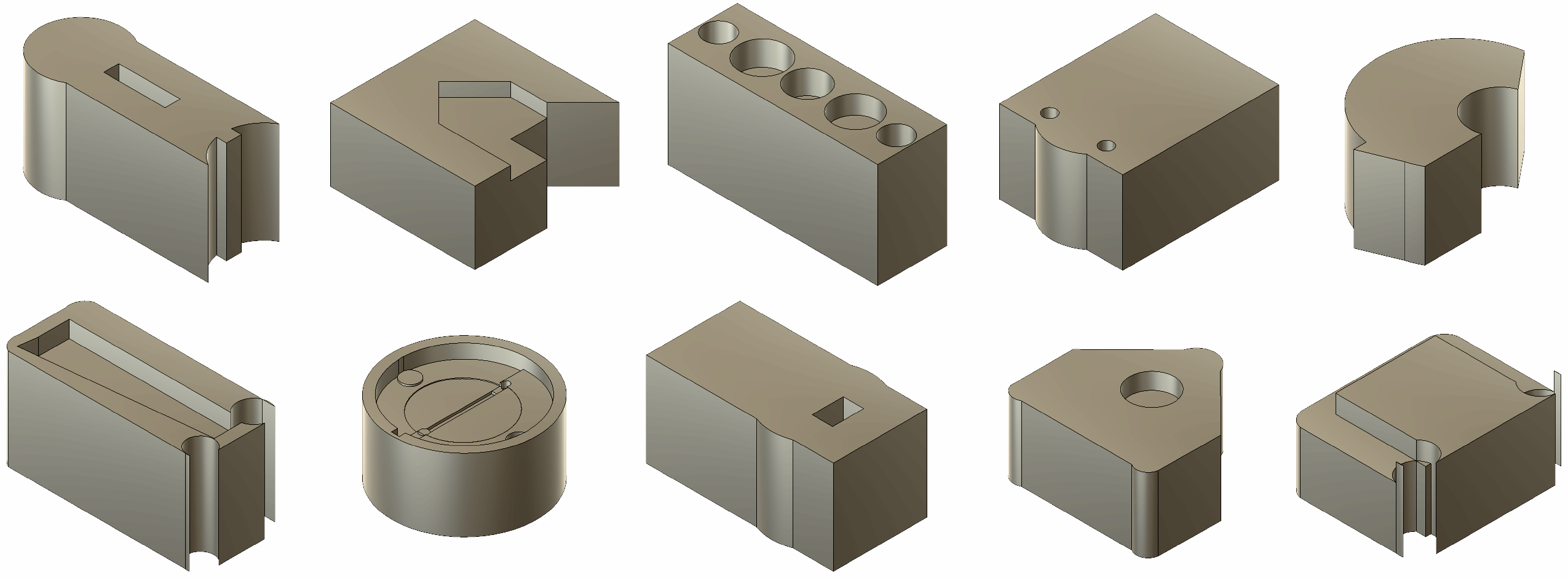} 
    \caption{Examples of 3D geometry produced by extruding the closed profiles of sketches generated from CurveGen.}
    \label{fig:sketch_gen_3d}
\end{figure*}

\subsection{Perceptual Evaluation}
To understand how engineering sketches generated by each model compare to human designed sketches, we perform a perceptual evaluation using human subjects. In our two-alternative forced choice study, each participant is presented with one human designed and one generated engineering sketch and asked: ``Which sketch is more realistic?". Brief instructions are provided, including an illustration of an engineering sketch used in context, similar to Figure~\ref{fig:teaser}. We evaluate 1000 unique generative sketches from each model, with a consistent set of 1000 human designed sketches from the SketchGraphs test set.
For each pair of sketches, we log the responses of three human subjects and use the majority answer. We conduct the study using workers from Amazon Mechanical Turk. Figure~\ref{fig:study_result}, left shows the percentage of generated sketches classed as more realistic than human designed sketches; higher values are better. A value of 50\% indicates the generated sketches are indistinguishable from human design. Figure~\ref{fig:study_result}, right shows the inter-rater agreement calculated as a percentage between each of the three human subjects. A lower value indicates there is more confusion between the generated and human designed sketches. The study results show that human subjects find CurveGen output to be the most realistic of the generated engineering sketches.

\begin{figure}[b]
    \centering
    \includegraphics[width=1\linewidth]{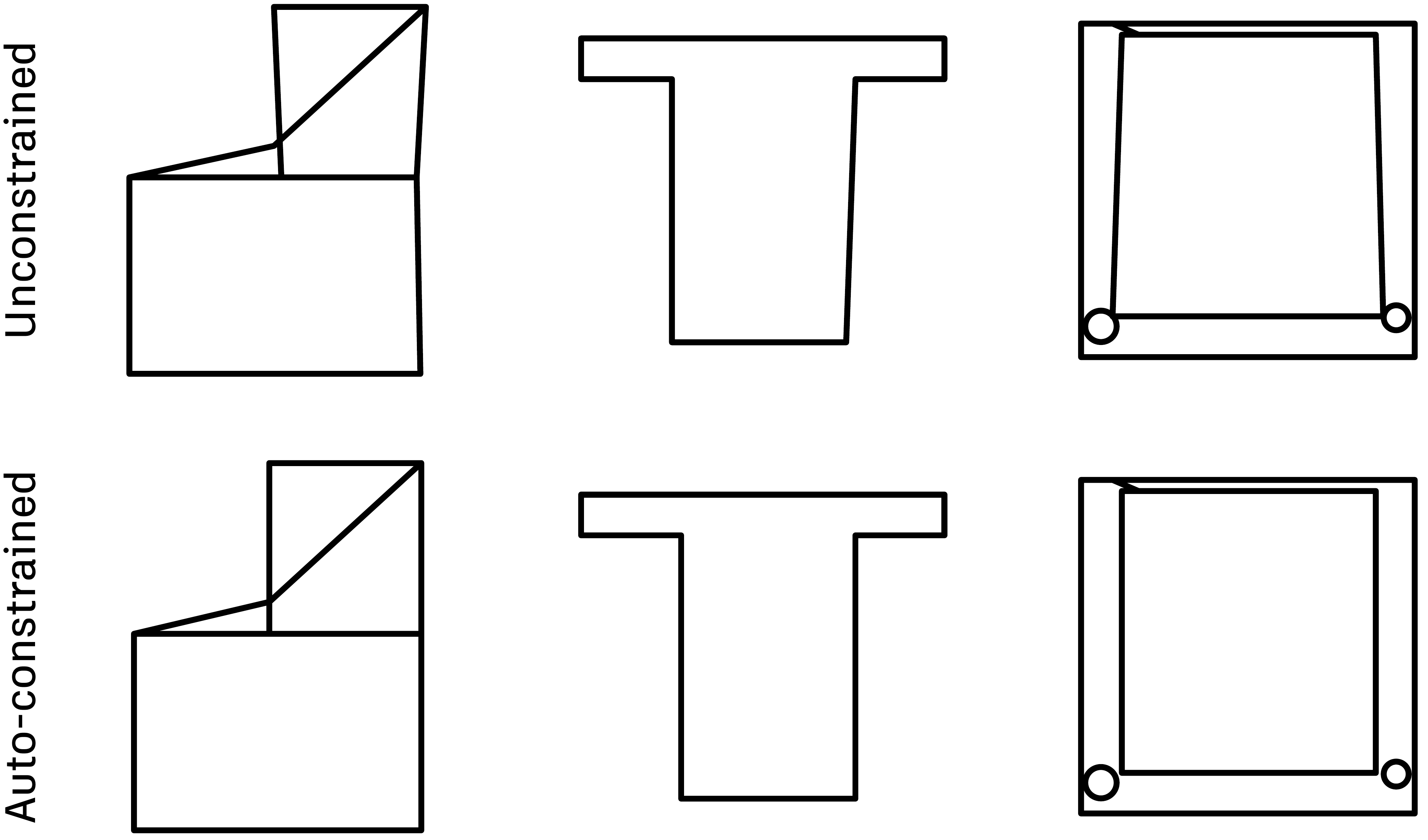} 
    \caption{Comparison of sketches generated by CurveGen before (top) and after (bottom) applying automatic constraints in Autodesk AutoCAD.}
    \label{fig:sketch_gen_qual_constraints}
\end{figure}

\subsection{Qualitative Results}
Figure~\ref{fig:sketch_gen_qual} shows human designed sketches from the SketchGraphs dataset beside randomly selected \textit{valid} sketches generated using the CurveGen, TurtleGen, and SketchGraphs generative models. 
We observe that CurveGen in particular is able to consistently produce sketches with closed loops, symmetrical features, perpendicular lines, and parallel lines. We provide additional qualitative results in Section~\ref{sec:sup_qual} of the supplementary material.

\paragraph{Sketch to Solid CAD Models}
A key motivation of the current work is to enable the synthesis and composition of solid CAD models. In Figure~\ref{fig:sketch_gen_3d} we demonstrate how engineering sketches generated by CurveGen with closed loop profiles can be lifted into 3D using the extrude modeling operation in a procedural manner.

\paragraph{Sketch Constraints}
The geometric output of our generative models can be post-processed to apply sketch constraints and build a constraint graph. Figure~\ref{fig:sketch_gen_qual_constraints} shows how the auto-constrain functionality in Autodesk AutoCAD can enforce parallel and perpendicular lines within a given tolerance. The unconstrained output from CurveGen is shown on the top row, and has a number of lines that are close to perpendicular. The auto-constrained output on the bottom row snaps these lines to perpendicular and establishes constraints for further editing.

\section{Conclusion}
\label{sec:conclusion}
In this paper we presented the CurveGen and TurtleGen generative models for the task of engineering sketch generation and demonstrated that they produce more realistic output when compared with the current state-of-the art. We believe engineering sketches are an important building block on the path to synthesis and composition of solid models with an editable parametric CAD history. 
Promising future directions include modeling higher-order constructs such as constraints and repetitions in the underlying design.

{\small
\bibliographystyle{ieee_fullname}
\bibliography{main}
}

\clearpage
\setcounter{section}{0}
\renewcommand\thesection{\Alph{section}}
\renewcommand\thesubsection{\thesection.\arabic{subsection}}

\section{Supplementary Material}

\subsection{TurtleGen Program Example}
\label{sec:sup_turtle_program}
One possible TurtleGen program for the sketch in Figure~\ref{fig:sketch_rep} is as follows: 
\[
\begin{array}[]{l}
 [\\
 \indent loopstart((86,43)),\\
 \indent line((169,0)),\\
 \indent line((0,170)),\\
 \indent line((-169,0)), \\
 \indent arc((-86, -85),(86, -85)), \\
 \indent loopstart((86,85)), \\
 \indent circle((43,43),(-43,43),(-43,-43)) \\
 ] \\
\end{array}
\]

The pen starts and returns to $(0,0)$ for each loop. The first five lines draw the outer loop, lifting the pen to $(86,43)$ and drawing counter clock-wise. The last two lines draw the inner circle, lifting the pen to $(86,85)$ vertex, and drawing counter clock-wise.

\subsection{Additional Qualitative Results}
\label{sec:sup_qual}
\begin{figure*}
    \centering
    \includegraphics[width=0.98\linewidth]{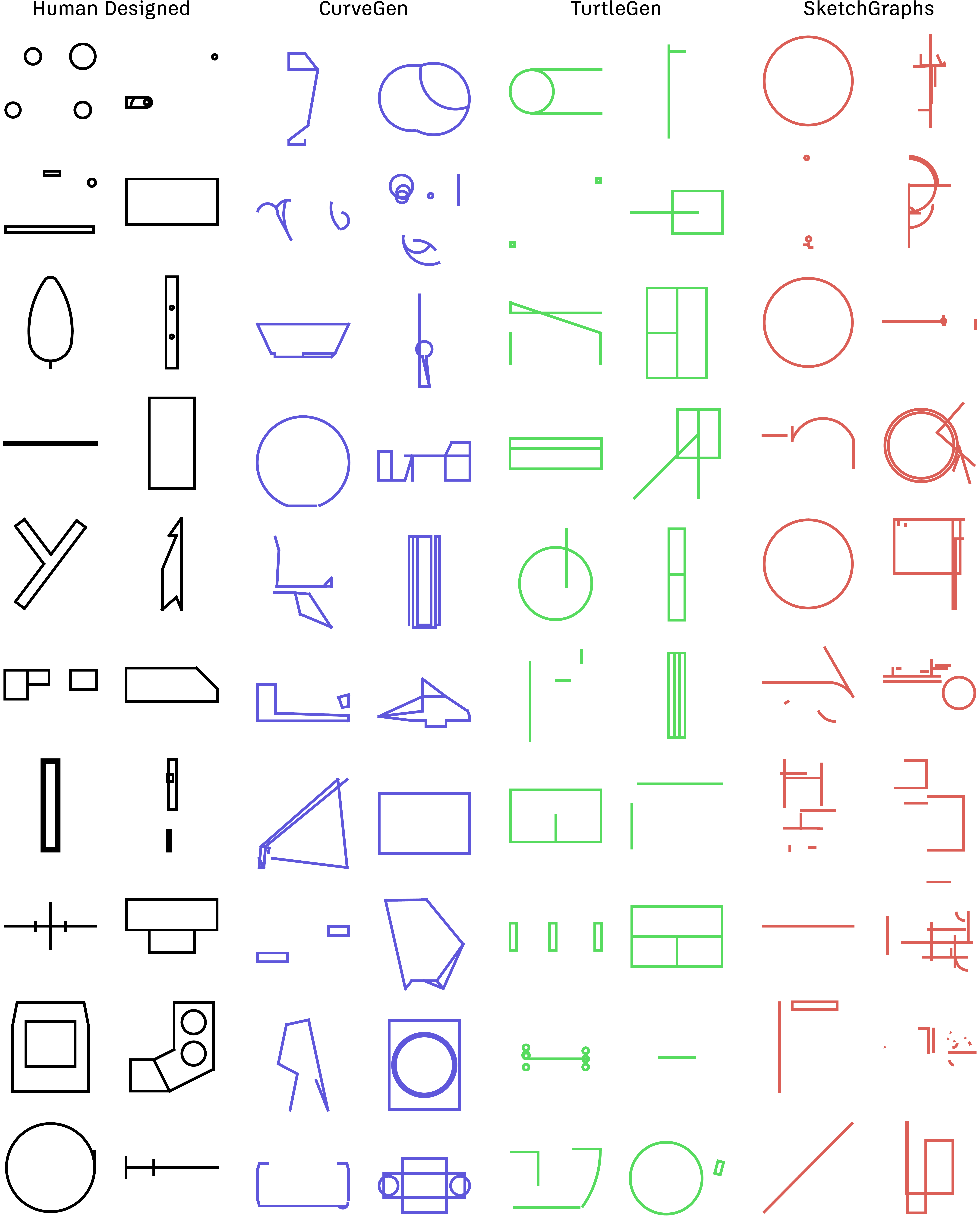} 
    \caption{Additional qualitative sketch generation results. From left to right: human designed sketches from the SketchGraphs dataset, randomly selected sketches generated using the CurveGen, TurtleGen, and SketchGraphs generative models.}
    \label{fig:sketch_gen_qual_sup1}
\end{figure*}

\begin{figure*}
    \centering
    \includegraphics[width=0.98\linewidth]{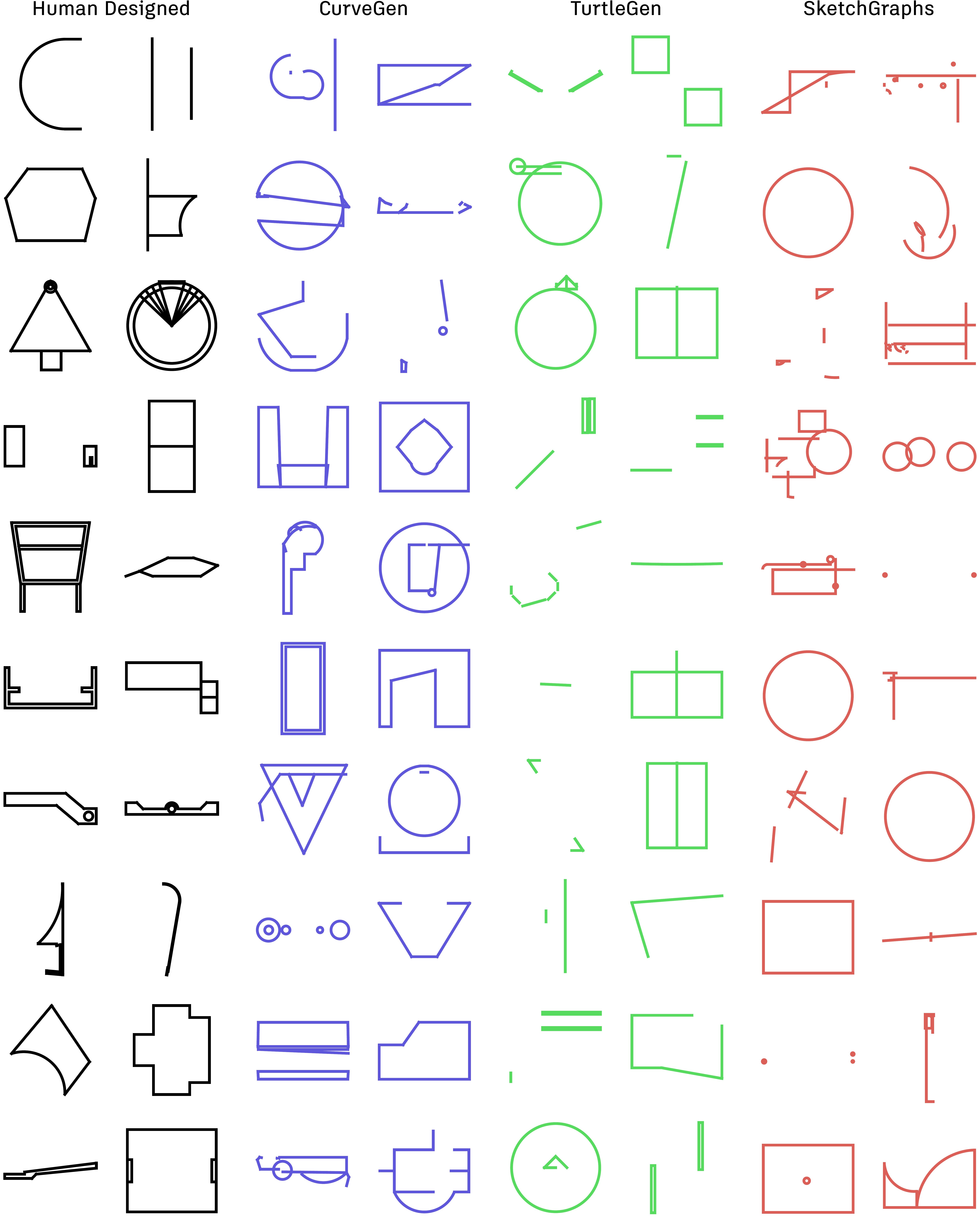} 
    \caption{Additional qualitative sketch generation results. From left to right: human designed sketches from the SketchGraphs dataset, randomly selected sketches generated using the CurveGen, TurtleGen, and SketchGraphs generative models.}
    \label{fig:sketch_gen_qual_sup2}
\end{figure*}

We provided additional qualitative sketch generation results in Figure~\ref{fig:sketch_gen_qual_sup1} and Figure~\ref{fig:sketch_gen_qual_sup2}.

\end{document}